\documentclass[runningheads]{llncs}
\usepackage{graphicx}
\usepackage{amsmath,amssymb} 
\usepackage{color}
\usepackage{subfig}

\usepackage{paralist} 
\usepackage{epsfig}
\usepackage{epstopdf}
\usepackage{amsmath}
\usepackage{amssymb}
\usepackage{multirow}
\usepackage{colortbl}
\usepackage{bm}
\usepackage{color}
\usepackage{floatrow}
\usepackage{booktabs}

\usepackage{comment}

\definecolor{mycyan}{cmyk}{.3,0,0,0}
\newcommand{\ie}{\textit{i}.\textit{e}. }

\newcommand{\etal}{\textit{et}~\textit{al.} } 

\begin{document}
\pagestyle{headings}
\mainmatter
\def\ECCVSubNumber{7270}  

\title{A Benchmark for Temporal Color Constancy}

\titlerunning{A Benchmark for Temporal Color Constancy}
\author{Yanlin Qian\inst{1,2,3} \and
Jani K\"apyl\"a\inst{1} \and
Joni-Kristian K\"am\"ar\"ainen\inst{1} \and 
Samu Koskinen \inst{2} \and
 Jiri Matas \inst{3}
}
\authorrunning{Qian et al.}
\institute{
Computing Sciences, Tampere University
\and
Huawei Tampere Research Center
\and
Center for Machine Perception, Czech Technical University in Prague
}

\maketitle
\begin{abstract}
Temporal Color Constancy (CC) is a recently
proposed approach that challenges the conventional single-frame color constancy. The conventional approach
is to use a single frame - shot frame - to estimate
the scene illumination color. In temporal CC, multiple frames from the view finder sequence are used to estimate the color.
However, there are no realistic large scale temporal
color constancy datasets for method evaluation. In this
work, a new temporal CC benchmark is introduced.
The benchmark comprises of 
(1) 600 real-world sequences recorded with a high-resolution mobile phone camera,
(2) a fixed train-test split which ensures consistent evaluation, and
(3) a baseline method which achieves high accuracy
in the new benchmark and the dataset used in previous
works. Results
for more than 20 well-known color constancy methods including
the recent state-of-the-arts are reported in our experiments.
\end{abstract}

\section{Introduction}
\label{sec:intro}

The human visual system perceives colors of objects independently of the incident illumination. This ability to perceive the colors in varying conditions as the scene is
viewed under a white light is known as color
constancy (CC)~\cite{finlayson2013corrected}.
To achieve this property, computational color constancy algorithms are used in Image Signal Processor (ISP) pipelines of digital cameras
to provide an estimate of the color of the illumination of the captured scene.

The existing color constancy algorithms can be mainly classified into two categories: 1) static methods and 2) learning-based methods.
Gijsenij~\etal~\cite{gijsenij2011computational} defined
a third class, gamut-based methods, in their survey. 
Since the gamut methods often require training examples to define
a target gamut~\cite{gijsenij2010generalized} we include
them to the learning-based category.
{\em Static methods} do not rely on training data, but
are based on assumed statistical or physical properties
of the image formation. For instance, Gray-world~\cite{buchsbaum1980spatial} relates the averaged pixel values to the global illumination and Gray pixel~\cite{yang2015efficient} and its extension~\cite{Qian_2019_CVPR} identify achromatic pixels using the properties of the lambertian model or dichromatic reflection model to reveal illumination, respectively.
{\em Learning-based methods} learn to map input image
features to the illumination estimate. Learning-based methods
can operate in the chroma space
(Corrected moments~\cite{finlayson2013corrected} and
Convolutional CC~\cite{barron2015convolutional}) or
in the spatial space full of rich semantic information
(FC4~\cite{hu2017cvpr}). Static methods are easier to
implement on commodity ISP hardware, but the recent
advantages in the mobile CPUs and GPUs have made it
intriguing to investigate whether the better performing
learning-based methods can replace static methods.

The above computational color constancy methods estimate
the illumination color from a single frame - referred to as the "shot frame" in our work. However, recently
Qian~\etal~\cite{yanlin2017iccv} proposed an
approach where multiple frames preceding the shot frame
are also used in the estimation - an approach that
can be termed as {\em temporal color constancy} or
multi-frame color constancy. 
They proposed
a recurrent network architecture based on AlexNet semantic
features and recursive network module for sequential processing.
The temporal color constancy is a realistic model of the process in a camera, where focus, gain, expose time and white balance  is constantly adjusted given a video stream that is displayed to the user, until the "shoot" button is pressed. 
The
experiments were conducted on the
SFU Gray Ball dataset~\cite{ciurea2003large} that
is captured with a video camera
where a calibration target is visible in every frame.
Qian~\etal demonstrated superior accuracy for
the temporal multi-frame setting vs. the conventional single-frame setting, 
but it is unclear to which extend the SFU Gray Ball video clips are related to real use cases of customer photography.
SFU Gray Ball consists of 15 sequences,
the sequences are captured over long time duration
and physically distant locations, 
and the frame
resolution is low ($240\times 320$). Moreover, the ground truth visible in
every frame can convey unintentional cues to deep net methods
even if masked.

\vspace{\medskipamount}
\noindent Our work makes the following contributions:
\begin{compactitem}
\item We release a \textbf{temporal color constancy (TCC) benchmark}. The dataset consists of 600 sequences of varying length (from 3 to 17 frames).
The dataset covers indoor and outdoor scenes with varying weather and daylight conditions, and is till now the largest realistic temporal dataset.

\item We make a \textbf{benchmark analysis} with over 20 statistical and learning-based single and the existing temporal methods, using a fixed train-test setting. 
\item We propose a strong \textbf{temporal color constancy baseline}, termed as TCC-Net, that achieves state-of-the-art results on the new dataset and the previously used SFU Gray Ball, with fast inference speed and light memory footprint. 
\end{compactitem}
TCC-benchmark and TCC-Net
will be made publicly available as an open-source project, to facilitate fair comparison and development of novel temporal color constancy ideas. We also provide wrapper functionality for experimenting with other datasets such
as the NUS dataset~\cite{Cheng14} and include implementations
of the recent methods such as FC4~\cite{hu2017cvpr} and C4~\cite{yanlin2017iccv}.

\section{Related work}
\label{sec:related}

\paragraph{Computational color constancy (CC)}
refers to the algorithms that estimate the illuminant color from
an image. Gijsenij~\etal\cite{gijsenij2011computational} provide a comprehensive survey of the contemporary methods and divide them under three categories: i) static, ii) gamut-based
and iii) learning-based methods. 
The static methods do not require training data.
Well-known static methods and commonly used baselines are
Gray-world~\cite{buchsbaum1980spatial} and
General Gray-world (inc. multiple variants)~\cite{van2007using}.
More recent static methods are Gray Pixel~\cite{yang2015efficient} and "Grayness Index"~\cite{Qian_2019_CVPR}. The static methods are
inferior in the single dataset setting where training and test images are drawn from the same dataset, but outperform learning-based methods in the cross-dataset evaluations~\cite{Qian_2019_CVPR}. In Gijsenij's taxonomy the gamut-based methods operate in the color spaces and thus
omit the spatial domain information. A strong baseline is Gamut Mapping~\cite{gijsenij2010generalized}. 
In our work, we assign the gamut-based methods to the
learning-based methods if they use training data
such as~\cite{gijsenij2010generalized}.
More recent methods operating in the colour spaces
are Corrected moments~\cite{finlayson2013corrected},
Convolutional CC~\cite{barron2015convolutional} and its
Fast Fourier implementation (FFCC)~\cite{barron2017fourier}. The most recent
learning-based methods are based on deep architectures that use pre-trained backbone networks to extract rich semantic features:
FC4~\cite{hu2017cvpr} and C4~\cite{yu2020aaai}.
We include the mentioned methods to our experiments
since they report state-of-the-art results
for various single-frame datasets.

\paragraph{Temporal color constancy}
has received less attention than the single frame CC. Attention has been
paid on several special cases. 
For example, Yang~\etal~\cite{yang2011uniform} extract illuminant color from two distinct frames of a scene that contains specular surfaces (highlights). 
Prinet~\etal~\cite{prinet2013illuminant} propose a probabilistic and more robust version of the
Yang~\etal method. Wang~\etal~\cite{wang2011video} compute color constancy for video frames. In their approach existing
CC methods can be used and illuminant is estimated from multiple frames of a same scene where scene boundaries are automatically detected.
Yoo~\etal~\cite{yoo2019dichromatic} propose a color
constancy algorithm for AC bulb illuminated (indoor) scenes using a high-speed camera and Qian~\etal~\cite{Qian-2019-icip}
for a pair of images with and without flash.
However, the seminal work of temporal color constancy is Qian~\etal~\cite{yanlin2017iccv} who seeded the term and
proposed a temporal CC algorithm using semantic AlexNet features and a Long Short Term Memory (LSTM) recurrent neural network to process sequential input frames.
Qian~\etal method and the dataset used in their experiments are included to our experiments.

\paragraph{Public datasets} are available for the evaluation
of single-frame color constancy methods, for example,
Gehler-Shi Color Checker~\cite{gehler2008bayesian,shi2010re},
SFU Gray Ball~\cite{ciurea2003large} and
NUS~\cite{Cheng14}.\footnote{See \url{http://colorconstancy.com} for download links of datasets and methods.}
SFU Gray Ball is collected with a video camera and is therefore suitable for multi-frame color constancy experiments
\cite{yanlin2017iccv}. 
However, the SFU Gray Ball has very low resolution ($240\times320$), contains only 15 sequences, and
its capture procedure does not correspond to the consumer
still photography. 
Yoo~\etal\cite{yoo2019dichromatic} have published the dataset of 80 sequences used in their experiments, but
their sequences were specifically designed for AC bulb illumination experiments and high-speed capturing.
Prinet~\etal\cite{prinet2013illuminant}
released a small dataset of 11 sequences used in their video color constancy experiments. 
In summary, the existing multi-frame color constancy datasets are small
and ill-suited for generic consumer still photography color constancy studies. Therefore we introduce a new dataset of 600 sequences
captured with a rooted mobile phone that makes the multi-frame
capture invisible to the mobile phone user and therefore better resembles consumer still photography.

\section{Dataset}
\label{sec:dataset}

The multi-frame temporal color constancy (TCC) dataset was collected by university students who captured the
shots during their free time. They were not given instructions but asked to take photographs whenever they wish. Students were
given a Huawei Mate 20 Pro mobile phone which is one of the high-end models and was rooted
and re-programmed to automatically start storing raw sensor images when the camera application was launched.
The sensor images were linked to the shot frames using
the date and time tags of the files.

\subsection{Image Capture}

\begin{figure}[t]
\centering
\includegraphics[width=0.9\linewidth]{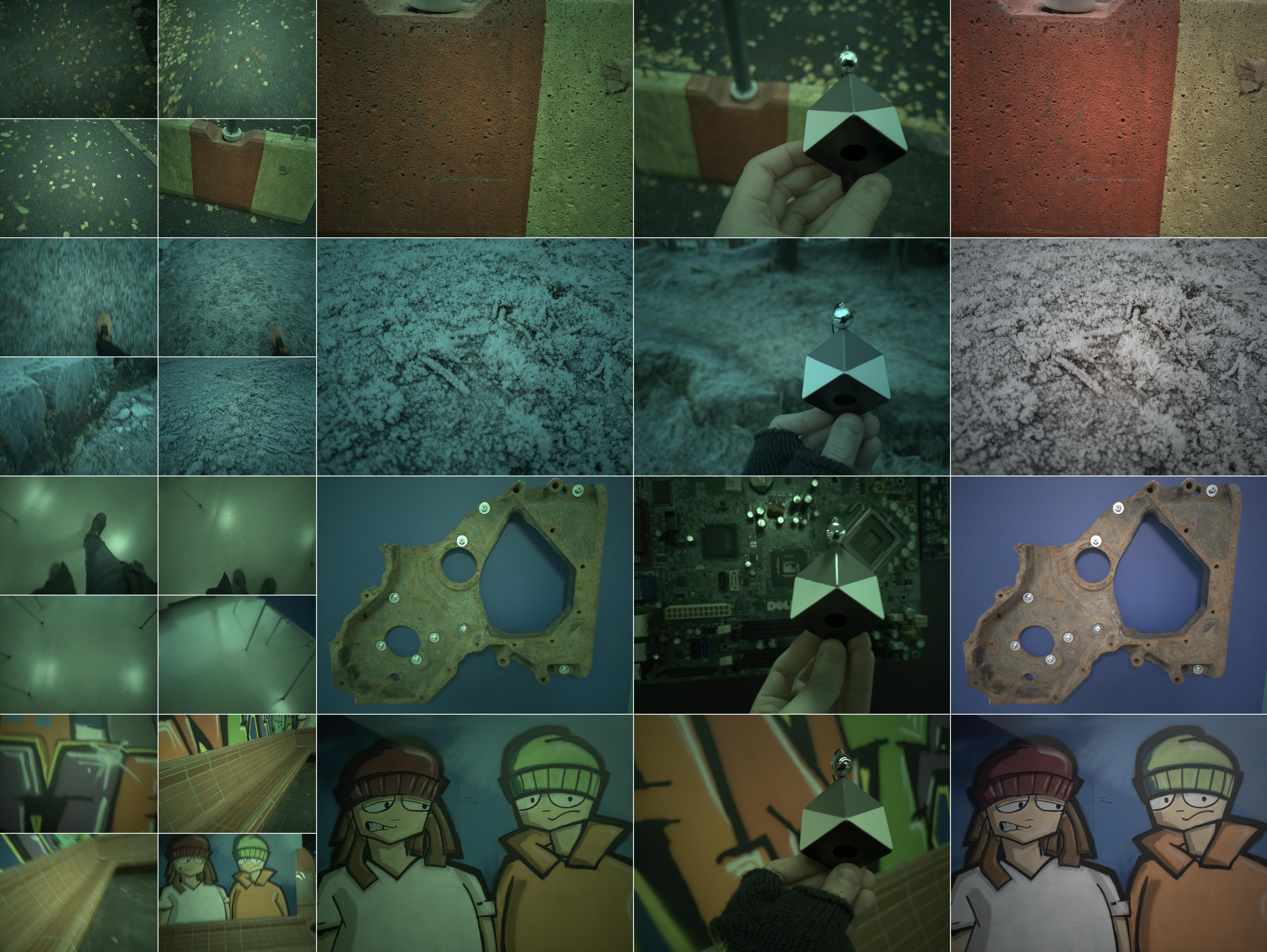}\\
\includegraphics[width=0.9\linewidth]{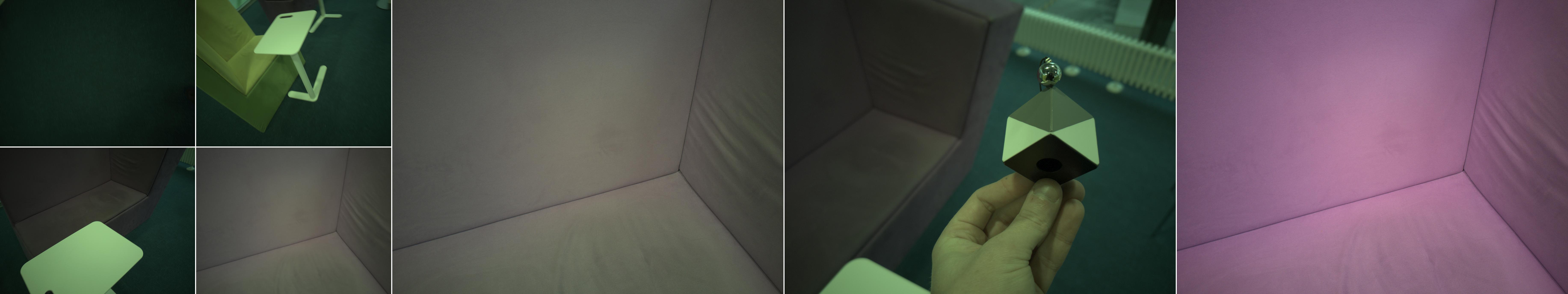}\\
\caption{Examples of 5 frame sequences in the collected TCC dataset. From each sequence there
are (left-to-right): 4 viewfinder frames, the shot frame, the calibration target frame and the color corrected shot frame. Note that sensor specific color correction is not applied, only color constancy. Gamma correction (2.2) is applied for better visualization. 
}
\label{figure:shotframe}
\end{figure}

The rooted phone saves the frames as
unprocessed 16-bit 3648$\times$2736 Bayer pattern images. High-resolution frame transfer from the ISP memory to the mobile phone storage memory requires special functionality that limits the practical capture frame rate to 1-3 frames per second (FPS).

To resolve the illuminant color ground truth the shot
frame scenes need to be captured with a color calibration
target installed into the scene. For example,
in the Gehler-Shi dataset there is a Macbeth
color checker calibration target visible in the images. In the SFU Gray Ball dataset a gray ball calibration target is attached to the video camera and is therefore visible in all captured frames.
In our dataset we wanted to avoid using visible targets
since they may unintentionally convey information to
the learning-based methods even if they are masked in
the training and test sets.

Similar to SFU Gray Ball we used a gray surface calibration target, SpyderCube (Figure~\ref{figure:shotframe}), which is put into
the shot scene instantly after the shot.\footnote{The target is always at the image center so that color shading has minimal effect on the ground truth.} The students were instructed to take one shot of the calibration
target in the location which was the main target or location
in their photograph. 
The captured sequences contain 3-17 frames depending
on the viewfinder duration (Figure~\ref{figure:shotframe}). The SpyderCube object
contains two neutral 18\% gray surfaces, from which
the one that better reflects the casting illumination
was annotated and used to
compute the ground truth illumination color. The
ground truth was verified by manually checking all
sequences using the ground truth correction. In total, 600 sequences were recorded and verified during different times
of day, in various indoor and outdoor locations and in
various weathers during the time period of October 2019 to January 2020.

In the dataset project page we provide linear demosaiced images in the PNG format with the pixel values
normalized to $[0,1]$, with a black level of zero and with no saturated pixels. The format correspond to that of Gehler-Shi dataset which is a popular evaluation set in color constancy literature. The black level of the specific camera sensor and device is $256$ and the saturation level is at $4095$. The final RGB images are of the
resolution 1824$\times$1368.

\subsection{Dataset Statistics and Performance Metrics}

\begin{figure}[t]
\centering
\includegraphics[width=0.40\linewidth]{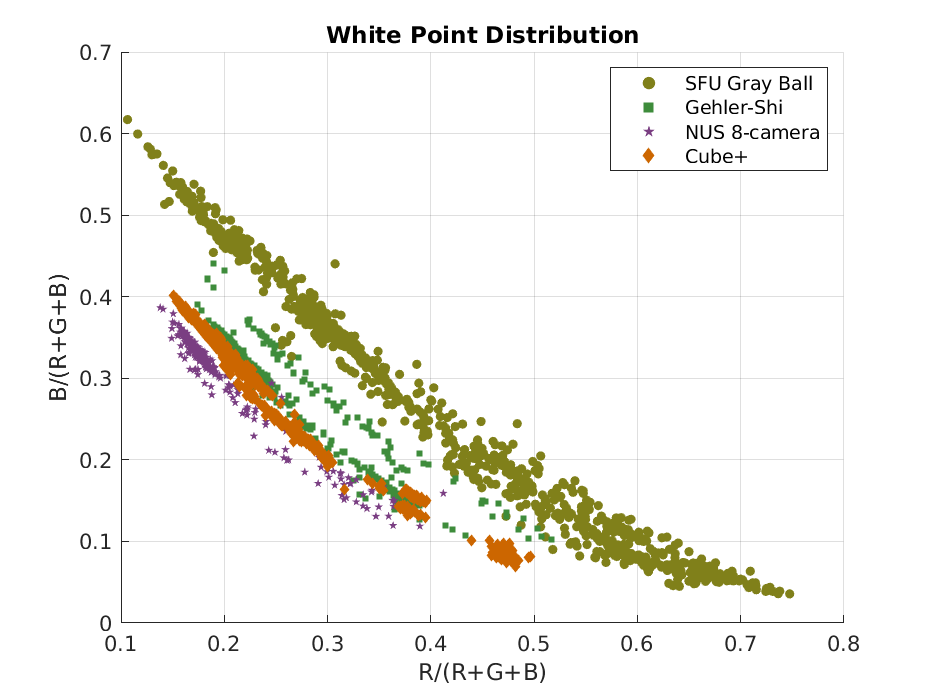}
\includegraphics[width=0.40\linewidth]{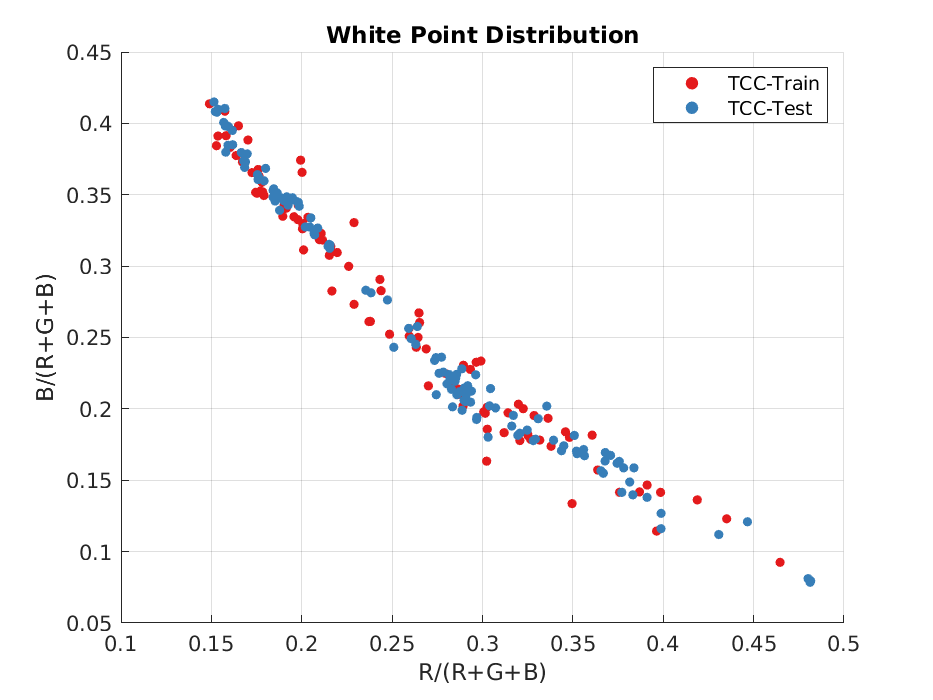} 
\\
\includegraphics[width=0.40\linewidth]{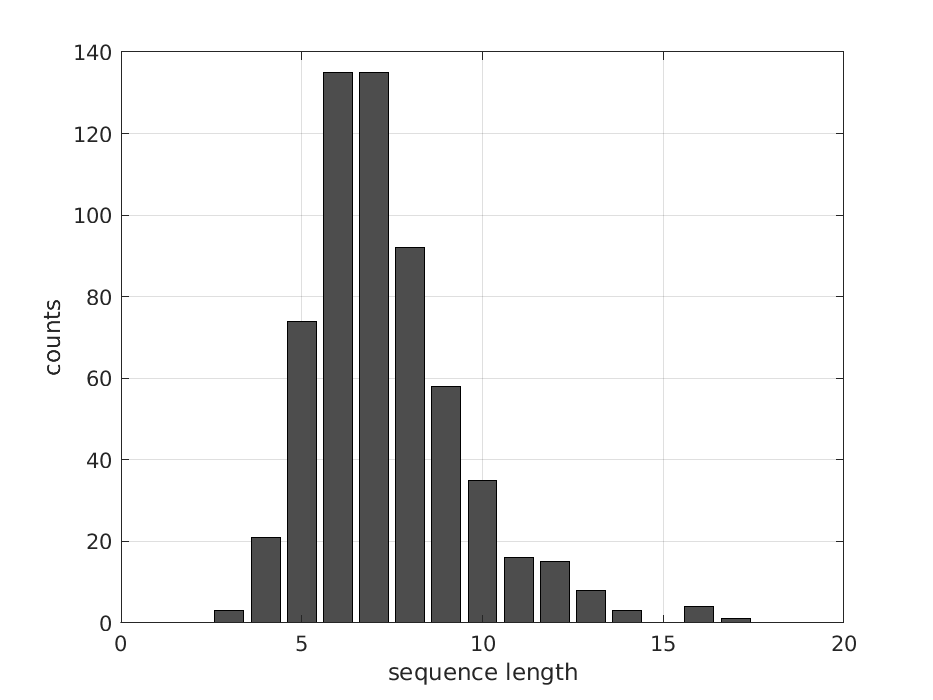} 
\includegraphics[width=0.40\linewidth]{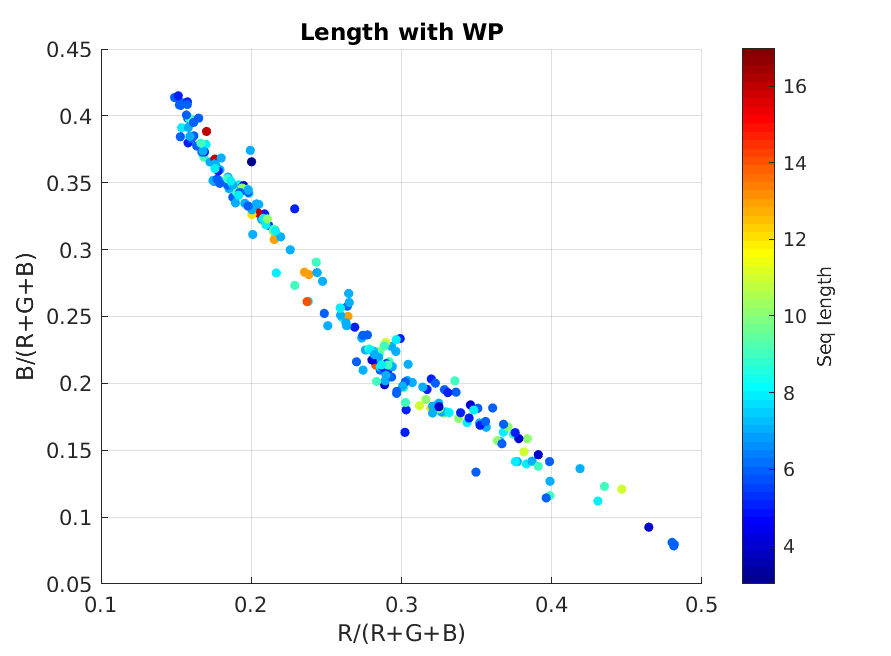} 
\caption{Top-left: White point distributions of several popular benchmarks. Top-right: White point distribution of our new dataset. Bottom-left: Histogram of sequence lengths of TCC. Bottom-right: Correlation between sequence chromaticity (white points) and the sequence length.}
\label{fig:distribution_histogram}
\end{figure}

We profile the distributions of ground truth chromaticity values of several mainstream color constancy benchmarks (Gehler-Shi, NUS 8-camera, Cube+ and SFU Gray Ball datasets) in the top-left inset of Figure~\ref{fig:distribution_histogram}, while we show that of the new Temporal Color Constancy dataset (TCC benchmark) in the top-right position. Our chromacity
distributions are similar to the popular Gehler-Shi,
NUS and Cube+ datasets.\footnote{Note that SFU Gray Ball distribution is larger than others since the data was captured with a high-end Sony VX-2000 video camera that has separate sensors for each color channel and therefore less spectral cross-talk and better channel separation.}
The small spatial shifts between the datasets are mainly due to different cameras used in the datasets.

In the bottom left of Figure~\ref{fig:distribution_histogram} we draw the histogram of sequence lengths in the TCC benchmark. The mean length is $7.3$, median $7.0$ and mode is $8.5$.  The bottom-right inset of Figure~\ref{fig:distribution_histogram} shows the correlation coefficient between the sequence lengths and the ground truth vectors, which indicates that there is
no clear correlation between the sequence length and the global illumination.

\paragraph{The main performance measure}
\label{sec:perf_measure}
in our work is the {\em angular error} which is used in the prior
works~\cite{barron2017fourier,yanlin2017iccv}.
The angular error $\varepsilon$ is computed from
the estimated tri-stimulus (RGB) illumination vector $\hat{{c}}$ and the ground truth vector ${c}_{gt}$ as
\begin{equation}
  \varepsilon_{\hat{{c}},{c}_{gt}} = \arccos\left(\frac{\hat{{c}}\cdot {c}_{gt}}{\parallel \hat{{c}}\parallel  \parallel {c}_{gt}\parallel}\right) \enspace , 
  \label{eq:metric}
\end{equation}
where $\cdot$ denotes the inner product between the two vectors and $\parallel\parallel$ is the Euclidean norm.
As overall performance measures we report
{\em mean}, {\em median} and
{\em trimean}. Tukey's trimean is a measure of a probability distribution's location defined as a weighted average of the distribution's median and its two quartiles. In addition, we report the top quartile
(25\%), the worst quartile (worst 25\%) and the
worst 5\% numbers.

\section{Methods}
\label{sec:methods}

\subsection{Extensions of Single-frame Methods}
\label{sec:extensions}
The conventional single-frame methods are designed to
estimate the illuminant color from a single image -
the shot frame. However, it is straightforward to
extend the single-frame methods to the multi-frame setting.
A single-frame method is executed on every frame and
the per frame estimates are combined using a suitable statistical tool such as the {\em moving average}.
In the following we introduce temporal extensions of
the SoTA statistical and learning-based methods.

\paragraph{Temporal Grayness Index (T.GI):}
Qian~\etal\cite{Qian_2019_CVPR} proposed a substantial
extension of the Gray Pixel method of Yang~\etal\cite{yang2015efficient}. They introduced
{\em Grayness Index (GI)} that provides a spatial
grayness map of the input image and the pixels of
the highest gray index are selected for the illumination
estimation. In the temporal extension of GI,
T.GI, all frames over the time are combined to form a
multi-frame GI map from which the best pixels are
selected. 

\paragraph{Temporal Fast Fourier Color Constancy (T.FFCC):} 
We use the official temporal smoothing implementation released by the author of FFCC~\cite{barron2017fourier}. It is based on a 
simplified Kalman filter with a simplified transition model, no control model and varying observation noise. The current estimate (modeled as an isotropic Gaussian) is smoothed by multiplying with last observed estimate. For more details, we refer to the temporal smoothing section in~\cite{barron2017fourier}.

\subsection{Temporal Color Constancy Network}
\label{sec:our_method}

\begin{figure}[!t]
\floatbox[{\capbeside\thisfloatsetup{capbesideposition={left},capbesidewidth=5.5cm}}]{figure}[\FBwidth]
{
\begin{tabular}{cc}
\centering
Input: $I_1,,,I_\text{len}$ & Input: $\hat{I}_1,,,\hat{I}_\text{len}$ \\
\noindent\rule{2cm}{.8pt} & \noindent\rule{2cm}{.8pt} \\
Backbone-3-512 & Backbone-3-512 \\
2DLSTM-512-128 & 2DLSTM-512-128 \\
\end{tabular} 
\linebreak
\begin{tabular}{c}
concatenation \\
MaxPool2d \\
Conv-256-64 \\
Sigmoid \\
Conv-64-3    \\
Sigmoid	 \\
\noindent\rule{2.5cm}{.8pt} \\
Output: $\mathbf{y}$ \\
\end{tabular}
}
{\caption[]{The architecture of TCC-Net. 
``LayerName-$x$-$y$'' denotes a 2D layer of $y$ filters of the size $x \times x$ where the layer
is either a standard convolution layer,
a backbone network (e.g. SqueezeNet) or
a 2D LSTM. ``len'' denotes the length of the
input sequence where the shot frame is
$I_{len}$. From the shot frame, a pseudo
sequence of the same length is generated
using the procedure in~\cite{yanlin2017iccv}.
$\mathbf{y}$ is the illumination color
vector after the last sigmoid layer.}
\label{fig:tcc_architecture}}
\end{figure}

\begin{figure}[t]
    \centering
    \includegraphics[width=0.90\linewidth]{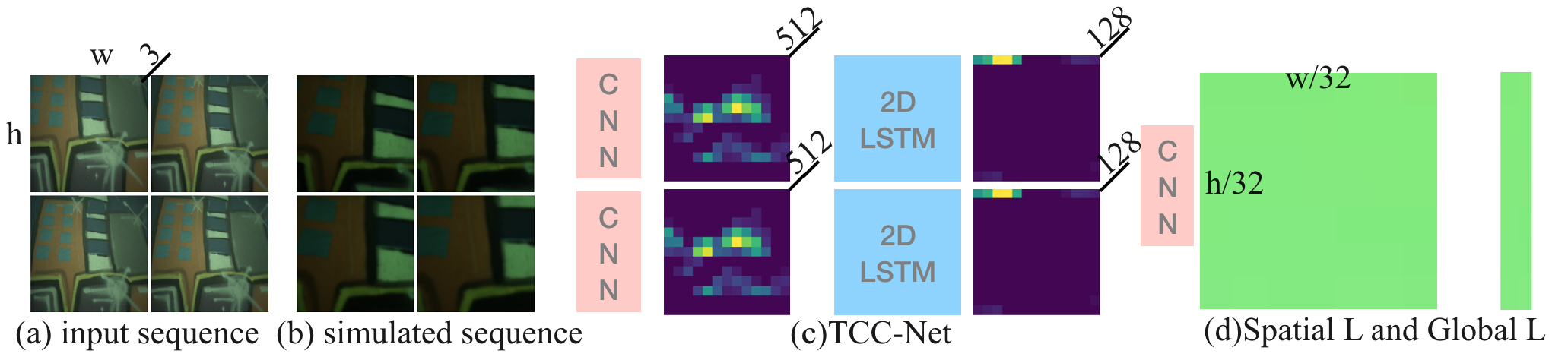}
    \caption{An overview of the TCC-Net processing pipeline:
    (a) input frame sequence;
    (b) a pseudo zoom-out sequence generated from
    the last (shot) frame;
    (c) from the both sequences the backbone
    network extracts 512-channel semantic features
    that are recursively processed by the
    2D LSTMs that output 128-channel features;
    (d) LSTM outputs are concatenated channel-wise
    and processed by a $1\times1$ convolution filter
    that produces a spatial illumination map.
    The global illumination vector $\mathbf{y}$ is calculated
    by average pooling.
    }
    \label{fig:pipeline}
\end{figure}

In the following, we propose a strong baseline for temporal color
constancy. The baseline is a deep network
architecture (TCC-Net) inspired by the RCC-Net
in~\cite{yanlin2017iccv}, but with the following
significant improvements:
1) a more powerful backbone network
for the semantic feature extraction,
2) 2D LSTM that provides
more effective spatial recurrent information and
3) support for variable length sequences. The overall
architecture is described in Fig.~\ref{fig:tcc_architecture}. TCC-Net adopts the
two CNN+LSTM branch structure from RCC-Net. The first branch,
the temporal branch, processes the image sequence,
and the second branch, the shot frame branch, processes a pseudo zoom-out sequence in the shot frame. In TCC-Net the both branches are based on
a novel 2D LSTM that produces spatio-temporal
information which are merged into a single RGB vector
at the end of the processing pipeline.

The backbone feature extraction network of RCC-Net (VGG-Net or AlexNet) is replaced with SqueezeNet~\cite{iandola2016squeezenet} in TCC-Net.
In a recent architecture for computational color constancy, FC4~\cite{hu2017cvpr}, the SqueezeNet~\cite{iandola2016squeezenet} was found
superior and this was verified by
our experiments (see Section~\ref{sec:experiments_ablation}).
Following \cite{hu2017cvpr}, we keep all layers up to the last convolution layer of SqueezeNet which outputs a 512-channel 2D feature map.

The second improvement is to adopt a 2D LSTM to temporally process sequences and learn a 2D spatial-temporal illumination feature map. We refer to the ordinary LSTM used by RCC-Net as ``1DLSTM'' due to the fact that its memory cells and the hidden states are encoded as 1D vectors. Although several 1DLSTMs can be stacked to learn more complex sequence-to-sequence mapping, the nature of 1DLSTM hinders its representative power for spatial information. 2DLSTM, introduced in \cite{xingjian2015convolutional}, extends 1DLSTM to
2D space by using convolutional structures in both input-to-state and state-to-state transitions.
Combining these changes, we have an end-to-end deep network which predict spatial illumination. 
To get the global estimate vector, averaging (or more advanced manipulation, \textit{e.g.} confidence weighted averaging in \cite{hu2017cvpr}) is applied.

TCC-Net provides native support to varying-length input. This is implemented by the dynamic computational graph feature supported in PyTorch.
In contrast, RCC-Net supports only a pre-defined
and fixed length sequences (3 or 5 frames in the original paper). With Nvidia GTX 1080ti the processing speed of TCC-Net is 6~ms per frame (only the network operations).

For better understanding of the network parameters
we present the key equations implemented in the
TCC-Net architecture.
For simplicity, the equations are given only for one branch, but the both branches share the similar stages.
Given an input sequence $\{I_1,\ldots ,I_\textit{len}\}$
and
the SqueezeNet backbone $F_{s}$
TCC-Net proceeds as
\begin{equation}
\begin{aligned}
& ~~~~\textit{Initialize the hidden state $H_0$ and the memory cell $C_0$ of 2D-LSTM} \\
& ~~~~\textit{for $t$ in range($1$,len)}: \\
&  ~~~~~~~~ \mathcal{X}_t = F_{s}(I_t) \\
& ~~~~~~~~ i_t = \sigma(W_{xi}\ast \mathcal{X}_t + W_{hi}\ast \mathcal{H}_{t-1} + W_{ci}\circ \mathcal{C}_{t-1} + b_i) \\
& ~~~~~~~~ f_t = \sigma(W_{xf}\ast \mathcal{X}_t + W_{hf}\ast \mathcal{H}_{t-1} + W_{cf}\circ \mathcal{C}_{t-1}+b_f) \\
& ~~~~~~~~ \mathcal{C}_t = f_t \circ \mathcal{C}_{t-1} + i_t \circ \tanh(W_{xc} \ast \mathcal{X}_t + W_{hc} \ast \mathcal{H}_{t-1}+b_c) \\
& ~~~~~~~~ o_t = \sigma(W_{xo}\ast \mathcal{X}_t + W_{ho}\ast \mathcal{H}_{t-1} + W_{co}\circ \mathcal{C}_{t}  +b_o) \\
& ~~~~~~~~ \mathcal{H}_t = o_t \circ \tanh(\mathcal{C}_t) \\
& ~~~~ L = F_{r}(\mathcal{H}_t )
\end{aligned}
\label{eq:tcc_net}
\end{equation}
where $i_t,f_t,o_t$ are 3D tensors and refer to the input, forget, and output gates of 2D-LSTM.
``$\ast$'' denotes convolution and ``$\circ$'' Hadamard product.  2D-LSTM has two parameters: the convolution kernel size $K$ (a larger value corresponds to faster illumination variations)
and the output channel size $H$ of the convolution
filter (corresponds to hidden channels of 1D-LSTM).
Ablation study of the both parameters is provided
in Section~\ref{sec:experiments}.
Figure~\ref{fig:pipeline} visualizes the workings of the TCC-Net pipeline.

\paragraph{Training:}
In all experiments we use the following settings.
The optimizer is
\textit{RMSprop}~\cite{hinton2012rmsprop} with the
learning rate $3e^{-5}$ and the batch size $1$. The network
was trained for 2,000 epochs. For data augmentation,
images were randomly rotated from $-30^\circ$ to
$+30^\circ$ and randomly cropped to the size
$[0.8,1.0]$ of the shorter size. Each patch was
horizontally flipped with the probability $0.5$.
The SqueezeNet backbone was initialized with the weights pretrained on ImageNet.
\section{Experiments}
\label{sec:experiments}
We run a large number of well-known methods on the new TCC Benchmark and report their accuracy in Section~\ref{sec:experiments_tcc}. 
In Section~\ref{sec:experiments_sfu} we verify good performance of the new baseline method (TCC-Net) with the previously used SFU Gray Ball dataset.
In Section~\ref{sec:experiments_ablation}
we provide ablation study of the main components
and parameters of TCC-Net.

\subsection{Method Comparison on TCC-benchmark}
\label{sec:experiments_tcc}

\begin{table}[t]
\newcommand{\tabincell}[2]{\begin{tabular}{@{}#1@{}}#2\end{tabular}}
\begin{center}
\caption{Method comparison with the TCC-benchmark. Performance metrics are based on the angular error (Section~\ref{sec:perf_measure}). The best results are
bolded and the second best underlined.
}
\label{tab:results}
\scriptsize
{
\begin{tabular}{l r r r r r r}
\toprule
{\bf Method} & {\em Mean} & {\em Med.} & {\em Tri.} &
{\em B25\%} &  {\em W25\%} & {\em W5\%} \\

\midrule
\multicolumn{2}{c}{{\em Single-frame Static}}\\
White-Patch  \cite{land1971lightness} & 11.20 & 10.42 & 10.87 & 1.87 & 21.48 & 26.20\\
Gray-World \cite{buchsbaum1980spatial}& 6.45 & 4.74 & 5.19 & 1.19 & 14.74 & 22.78 \\

Shades-of-Grey (p=4) \cite{finlayson2004shades} & 5.50 & 3.20 & 3.70 & 0.85 & 13.92 & 21.86 \\
General Grey-World (p=1 ,$\sigma$=9) \cite{van2007using} & 6.44 & 4.76 & 5.24 & 1.18 & 14.75 & 22.83 \\
1st-order Grey-Edge (p=1, $\sigma$=9) \cite{van2007using} & 5.46 & 4.09 & 4.25 & 1.01 & 12.84 & 21.06 \\
2nd-order Grey-Edge (p=1, $\sigma$=9) \cite{van2007using} & 5.10 & 3.62 & 3.85 & 1.00 & 12.00 & 20.48 \\
 PCA (Dark+Bright) \cite{Cheng14} & 5.45 & 3.00 & 3.68 & 0.96 & 13.78 & 22.93 \\
 Grayness Index (GI) \cite{Qian_2019_CVPR}  & 4.99 & 2.68 & 3.10 & 0.71 & 13.22 & 24.12 \\
 \multicolumn{2}{c}{{\em Temporal extensions}}\\
T.GI  & 4.73 & 2.96 & 3.39 & 0.82 & 11.38 & 17.42 \\

\midrule
\multicolumn{2}{c}{{\em Single-frame Learning-based}}\\
Pixel-based Gamut ($\sigma$=4) \cite{gijsenij2010generalized} & 6.90 & 5.53 & 6.20 & 1.18 & 14.72 & 19.19 \\
Edge-based Gamut ($\sigma$=3)  \cite{gijsenij2010generalized} & 8.69 & 7.58 & 8.12 & 2.00 & 17.16 & 20.54 \\
Intersection-based Gamut ($\sigma$=4) \cite{gijsenij2010generalized} & 8.46 & 7.94 & 7.85 & 2.03 & 16.60 & 20.80 \\
Natural Images Statistics \cite{gijsenij2011color}& 5.63 & 6.89 & 5.88 & 1.41 & 14.61 & 22.20 \\
LSRS \cite{gao2014eccv} & 6.61 & 4.92 & 5.52 & 1.67 & 13.90 & 21.37 \\
Exemplar-based Colour Constancy \cite{joze2014exemplar}& 5.24 & 3.88 & 4.21 & 1.38 & 11.58 & 19.82 \\
Chakrabarti \textit{et al.} 2015 ~\cite{chakrabarti2015color} Empirical & 4.26 & 2.60 & 2.82 & 0.51 & 11.07 & 16.43 \\
 Regression (SVR) \cite{funt2004estimating}& 4.00 & 3.09 & 3.45 & 1.36 & 7.81 & 11.07 \\
Bayesian \cite{gehler2008bayesian} & 4.25 & 2.86 & 3.16 & 0.93 & 9.97 & 16.27 \\
Random Forest \cite{cheng2015effective}& 3.76 & 2.66 & 2.94 & 0.74 & 8.54 & 13.14 \\
\text{AlexNet}-FC$^4$ \cite{hu2017cvpr}& 3.10 & 2.12 & 2.35 & 0.85 & 6.78 & 8.21 \\
\text{SqueezeNet}-FC$^4$ \cite{hu2017cvpr} & 2.84 & 2.10 & 2.23 & 0.74 & 6.39 & 7.83 \\
C4 (3 stage) \cite{yu2020aaai} & 2.37 & 1.60 & 1.76 & 0.57 & 5.58 & \underline{6.85} \\
FFCC(model Q) \cite{barron2017fourier}& \underline{2.33} & \underline{1.37} & \underline{1.60} & \underline{0.49} & 5.84 & 10.97 \\
\multicolumn{2}{c}{{\em Temporal extensions}}\\
T.FFCC  & 3.35 & 1.70 & 1.99 & 0.51 & 9.06 & 17.41 \\

\midrule
\multicolumn{2}{c}{{\em Temporal }}\\
RCC-Net \cite{yanlin2017iccv} & 2.74 & 2.23 & 2.39 & 0.75 & \underline{5.51} & 8.21 \\
Our (TCC-Net) & {\bf 1.99} & {\bf 1.21} & {\bf 1.46} & {\bf 0.30} & {\bf 4.84} & {\bf 6.34} \\
\bottomrule
\end{tabular}
}
\end{center}
\end{table}

The results for various single-frame static and learning-based methods (see the related work section), their temporal extensions (Section~\ref{sec:extensions}), the current temporal state-of-the-art (RCC-Net)~\cite{yanlin2017iccv}
and our temporal baseline (Section~\ref{sec:our_method})
are shown in Table~\ref{tab:results}. The results demonstrate that the recent deep learning based methods (FC$^4$ and C4) and the convolutional CC
(FFCC) are clearly superior to the conventional static and learning-based methods. These methods
improve the performance over the whole error distribution, \ie both the easy and difficult test
samples.
On our dataset the previous temporal state-of-the-art, RCC-Net~\cite{yanlin2017iccv}, is slightly inferior
to the best single-frame methods C4 and FFCC. 

The temporal extension of GI~\cite{Qian_2019_CVPR},
T.GI, improves its results. On the contrary,
T.FFCC, referred to as ``temporal smoothing''
in~\cite{barron2017fourier}, is inferior to its single-frame version. The Kalman filter extension of FFCC provides
smoother change of the illuminant estimates over
the frames, but the accuracy is worse than the non-smoothed estimates. We also test Prinet \etal\cite{prinet2013illuminant} and it achieves 7.51 mean error due to its assumption that the illumination remains constant over time. 

\begin{figure} [t]
    \centering
    \includegraphics[width=0.95\linewidth]{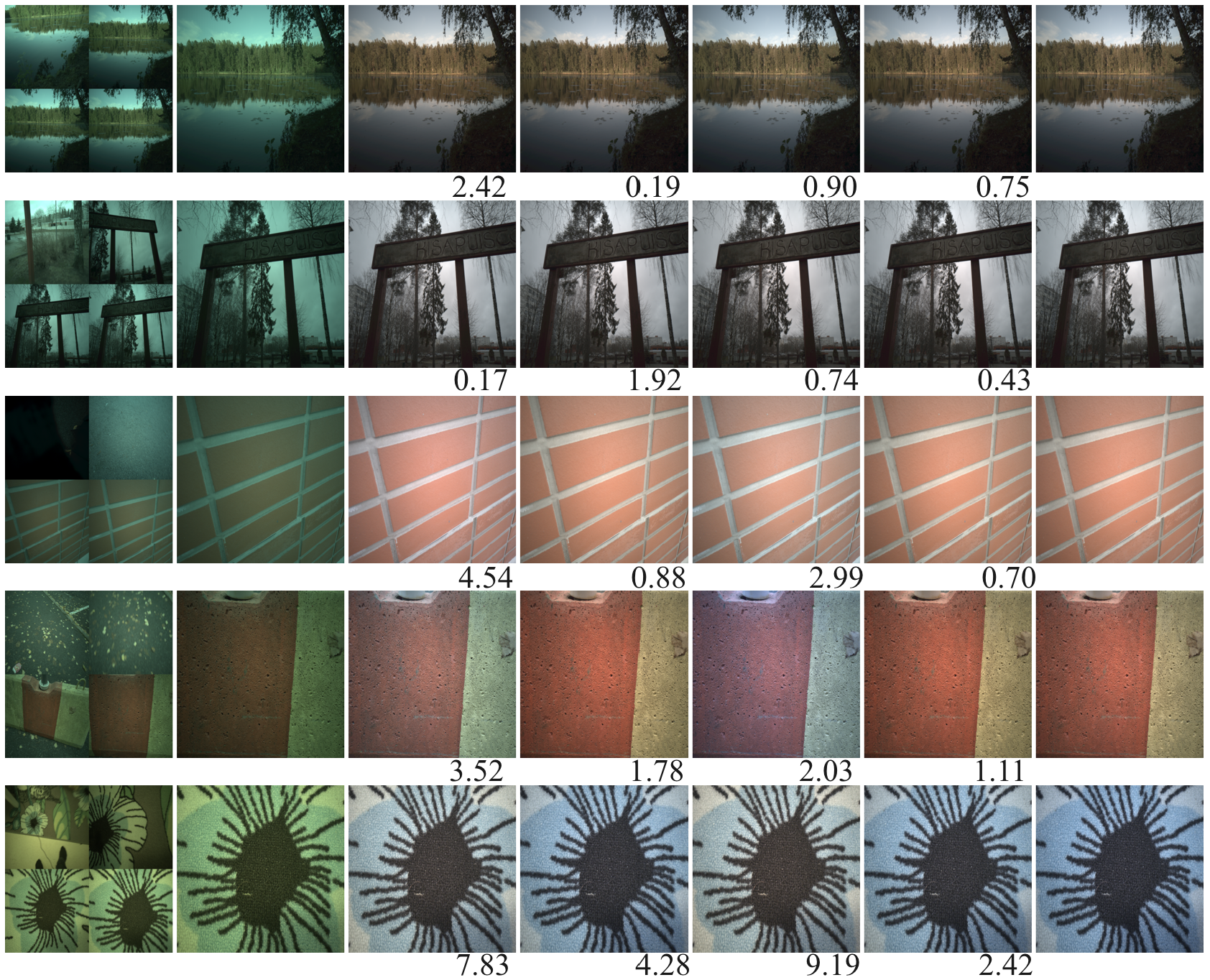}\\
    \caption{Color corrected TCC examples and their angular errors
    (left-to-right):
    1) four view finder frames;
    2) the shot frame;
    3) FC$^4$~\cite{hu2017cvpr};
    4) RCC-Net~\cite{yanlin2017iccv};
    5) FFCC~\cite{barron2017fourier};
    6) the proposed TCC-Net;
    7) ground truth correction.}
    \label{fig:qualitative_results}
\vspace{-3mm}
\end{figure}

The proposed TCC-Net (Model G in Table~\ref{tab:ablation}) obtains the best performance on all error measures and improves performance on the both easy and difficult cases. As compared to the previous state-of-the-art, RCC-Net, the performance improvement is over
35\% in the mean error and and 43\% in the median error.
Considering the fact that end-users are more sensitive to large estimation errors~\cite{Cheng14} and
$\le 3.0^\circ$ is generally considered as the sufficient accuracy, then W25\% error of the TCC-Net (4.84) is
closest to the practical use among all tested methods.

In Figure~\ref{fig:qualitative_results} are examples
of color-corrected images with various methods.
The first two examples demonstrate easy cases from
outdoors where all methods perform comparably well.
The third and fourth examples represent typical
view finder sequences toward a target which itself does not
provide visually-rich clue for inferring the illumination color.
In these sequences the two temporal methods, RCC-Net and
TCC-Net, provide the best results since they effectively exploit cues from the view finder frames. The last example is
a difficult case where the shot frame is a closeup of a tinted fabric material which can be of any plausible color. For the fifth sequence only the proposed TCC-Net provides an accurate estimate.

\subsection{Method Comparison on SFU Gray Ball}
\label{sec:experiments_sfu}
To validate the findings in the previous experiment
with the new TCC-benchmark, we replicated the
experiments in
Qian~\etal~\cite{yanlin2017iccv}, using their
metrics (the mean, median, worst 5\% and maximum errors)
and the SFU Gray Ball dataset. The results are collected
to Table~\ref{tab:grayball} (cf. Table~1 in \cite{yanlin2017iccv}).

On the temporal version of the SFU Gray Ball
dataset, the proposed TCC-Net again outperforms the RCC-Net~\cite{yanlin2017iccv},
with a clear margin. The difference of these two methods is particularly evident on the hardest cases as
TCC-Net obtains more than 40\% lower error on the
both worst 5\% and the maximum error metrics.
\begin{table}[t]
\newcommand{\tabincell}[2]{\begin{tabular}{@{}#1@{}}#2\end{tabular}}
\begin{center}
\caption{Method comparison with the SFU Gray Ball dataset (non-linear). The numbers for other methods are copied from the original papers and~\cite{yanlin2017iccv}.}
\label{tab:grayball}
\scriptsize
{
\begin{tabular}{l r r r r}
\toprule
{\bf Method} & {\it Mean} & {\it Med.} & {\it W5\%} & {\it Max}  \\
\midrule
\multicolumn{2}{c}{{\em Single-frame Static}}\\
Gray-World \cite{buchsbaum1980spatial}& 7.9 & 7.0 & -- & 48.1 \\
General Grey-World (p=1 (0),$\sigma$=9) \cite{van2007using} & 6.1 & 5.3 & -- & 41.2 \\
1st-order Grey-Edge (p=1, $\sigma$=9) \cite{van2007using} & 5.9 & 4.7 & -- & 41.2 \\
Gray Pixel \cite{yang2015efficient} & 6.2 & 4.6 & 20.8 & 33.3 \\
Shades-of-Gray \cite{finlayson2004shades} & 6.1 & 5.2 & -- & 41.2\\
 \midrule
\multicolumn{2}{c}{{\em Single-frame Learning-based}}\\
 Pixel-based Gamut ($\sigma$=5)  \cite{gijsenij2010generalized} & 7.1 & 5.8 & -- & 41.9 \\
 Edge-based Gamut ($\sigma$=3)  \cite{gijsenij2010generalized} & 6.8 & 5.8 & -- & 40.3 \\
Intersection-based Gamut ($\sigma$=9) \cite{gijsenij2010generalized} & 6.9 & 5.8 & -- & 41.9\\
Inverse-Intensity Chromaticity Space \cite{tan2008color} & 6.6 & 5.6 & -- & 76.2  \\
Random Forest \cite{cheng2015effective}& 6.1 & 4.8 & 13.1 & 30.6\\
LSRS \cite{gao2014eccv} & 6.0 & 5.1 & -- & -- \\
Natural Images Statistics \cite{gijsenij2011color}& 5.2 & 3.9 & -- & 44.5 \\
Exemplar-based Colour Constancy \cite{joze2014exemplar}& 4.4 & 3.4 & -- & 45.6 \\
ColorCat \cite{banic2014color} & 4.2 & 3.2 & -- & 43.7\\
\midrule 
\multicolumn{2}{c}{{\em Temporal}}\\
Prinet \textit{et al.}~\cite{prinet2013illuminant} & 5.4 & 4.6 & -- & -- \\
Wang \textit{et al.}~\cite{wang2011video} & 5.4 & 4.1 & -- & 26.8 \\
RCC-Net \cite{yanlin2017iccv} & \underline{4.0} & \underline{2.9} & \underline{12.2} & \underline{25.2}\\
Our (TCC-Net) & {\bf 2.8} & {\bf 2.3} & {\bf 7.1} & {\bf 13.9}  \\
\bottomrule
\end{tabular}
}
\end{center}
\vspace{-3mm}
\end{table}

\subsection{TCC-Net Ablation Study}
\label{sec:experiments_ablation}

Results with different components and parameter
settings of TCC-Net are given in Table~\ref{tab:ablation} and briefly discussed below.

\begin{figure}[t]
    \centering
    \includegraphics[width=0.95\linewidth]{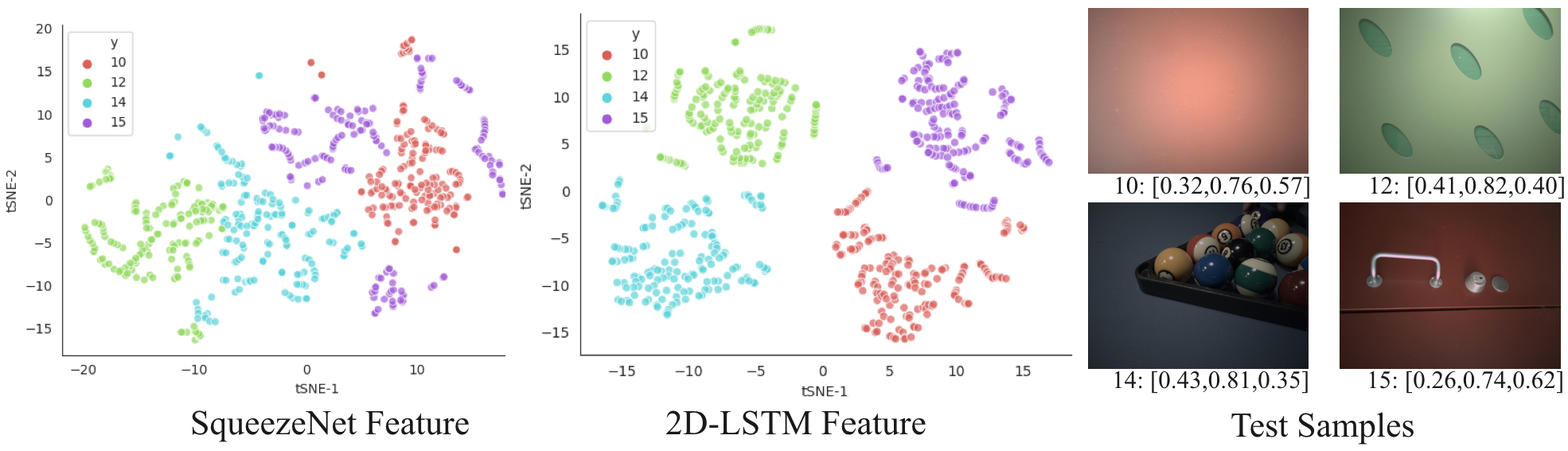}
    \caption{t-SNE visualizations of SqueezeNet and 2D-LSTM feature maps in the TCC-Net architecture. Colors represents different illuminations in the shot frames of the sequences \#10, \#12, \#14 and \#15 (on the right). Dots represent feature vectors (512 for SqueezeNet and 128 for 2D-LSTM) at different spatial locations of the shot frames.}
    \label{fig:tnse}
\end{figure}

\paragraph{Does LSTM help?}
The 1-branch TCC-Net (Model B in Table~\ref{tab:ablation}) without the LSTM module
becomes equivalent to SqueezeNet-FC$^4$ in Table~\ref{tab:results}. However, with the LSTM
module, for example the mean error is 11\% lower
than SqueezeNet-FC$^4$ which can be explained only
by the temporal information carried in the LSTM memory cell.
Additionally, Figure~\ref{fig:tnse} shows the t-SNE visualization \cite{maaten2008visualizing} of how LSTM representation is more discriminative than that of SqueezeNet backbone in our TCC-Net. t-SNE is used to visualize high-dimensional feature data. For each of the four selected samples shown in the right-hand-side of Figure~\ref{fig:tnse}, SqueezeNet backbone and 2D-LSTM output deep representations.
The representations are of the dimensions of (h,w,512) and (h,w,128), respectively, where  where h is the height, w width and 512 (or 128) the number of the feature channels. Contrast to the SqueezeNet backbone, LSTM exploits spatio-temporal information over multiple frames and provides
features which better represent the different illuminations.

\begin{table}[t]
\newcommand{\tabincell}[2]{\begin{tabular}{@{}#1@{}}#2\end{tabular}}
\begin{center}
\caption{Ablation study of TCC-Net with various different configurations. The default values for the number of LSTM channels is H=128 and for the convolutional kernel size K=5.
}
\label{tab:ablation}
\scriptsize
{
\begin{tabular}{l l r r r r r r r}
\hline
& {\bf TCC Configuration} & {\em Mean} & {\em Med.} & {\em Tri.} & {\em B25\%} &  {\em W25\%} & {\em W5\%} & {\em Mem. (MB)} \\
\hline
A & 2branch,AlexNet,1D-LSTM & 2.74 & 2.23 & 2.39 & 0.75 & 5.51 & 8.21 & 20.4\\
B & 1branch,SqueezeNet,1D-LSTM & 2.52 & 1.77 & 2.04 & 0.52 & 5.65 & 6.58 & 3.3 \\
C & 2branch,SqueezeNet,1D-LSTM & 2.20 & 1.55 & 1.65 & 0.43 & 5.05 & 6.18 & 6.6\\
D & 2branch,SqueezeNet,2D-LSTM,len1 & 3.27 & 3.46 & 3.32 & 2.07 & 4.44 & 4.80 & 68.8\\
E & 2branch,SqueezeNet,2D-LSTM,len5 & 2.50 & 1.78 & 1.99 & 0.53 & 5.65 & 6.94 & 68.8\\
F & 2branch,SqueezeNet,2D-LSTM(H=64) & 2.17 & 1.59 & 1.68 & 0.40 & 5.00 & 6.72 & 33.3\\
G & 2branch,SqueezeNet,2D-LSTM,(H=128) & 1.99 & 1.21 & 1.46 & 0.30 & 4.84 & 6.34 & 68.8\\
H & 2branch,SqueezeNet,2D-LSTM,(H=512) & 2.06 & 1.09 & 1.40 & 0.30 & 5.19 & 7.65 & 476.1\\
I & 2branch,SqueezeNet,2D-LSTM(K=1) & 2.01 & 1.42 & 1.58 & 0.34 & 4.65 & 5.48 & 11.0\\
J & 2branch,SqueezeNet,2D-LSTM(K=7) & 2.08 & 1.43 & 1.60 & 0.35 & 4.83 & 5.83 & 131.0\\

\hline
\end{tabular}
}
\end{center}
\vspace{-5mm}
\end{table}

\paragraph{Backbone network:}
The Model A in Table~\ref{tab:ablation} is the baseline as this configuration corresponds to RCC-Net in~\cite{yanlin2017iccv}. 
The effect of using
SqueezeNet instead of AlexNet backbone is evident
between the models A and C. The results with SqueezeNet
are superior to the results with AlexNet and
the memory footprint of SqueezeNet is substantially
smaller making it more practical for mobile
devices. Intriguingly, a single-branch TCC-Net without the pseudo sequence branch (Model B) also performs better than
the RCC-Net baseline (Model A) and thus verifies superior performance of SqueezeNet for color constancy.
By comparing Model B
and Model C it is clear that the two branch design
provides better performance than a single branch by a clear margin (the mean error is reduced by 12.7\%).

\paragraph{1D vs. 2D LSTM:} Model G is the main
model reported in Table~\ref{tab:results}.
The same
configuration but with 1D LSTMs is Model C. By comparing the performances of C and G
it is obvious that 2D LSTMs provide
better performance and achieve state-of-the-art in
the TCC and SFU Gray Ball benchmarks.
TCC-Net baseline (Model G) is a fully 2D convolutional
architecture that is the best found architecture for
illuminant estimation in temporal color constancy.

\paragraph{Dimensionality of LSTM hidden channels:}
Three different sizes of the LSTM
hidden channels, $H=\{64, 128, 512\}$, where tested
(Models F, G and H, respectively). 
For H=64 (Model F) the LSTM underfits and
for H=512 the network starts to overfit thus
making H=128 a good trade-off between training error and model generalization.

\paragraph{Kernel Size of 2D LSTM:}
The kernel size defines the amount of spatial correlations retained by the 2D LSTM. Kernel size K=1 means that the neighbor pixels do not affect to the LSTM inference. Different
kernel sizes were tested (Models G, I an J) and
the best results were achieved with K=5.

\paragraph{Varying-length Input:} One significant difference
to the previous state-of-the-art (RCC-Net)~\cite{yanlin2017iccv} is that TCC-Net allows an arbitrary number of input frames before the shot frame. We experimented on two fixed lengths, 1 (only the shot frame) and 5, and the arbitrary length
(Models D, E and G, respectively). The single-frame
results are the worst, five frames is the second best,
and arbitrary length achieves the best performance and
is the most convenient for the end-user cases where the
length of a view finder sequence is unknown.

\paragraph{Memory Footprint :} From the perspective of deploying the deep net into a GPU/NPU-supported consumer mobile platform, we profiled the memory footprints of
all TCC-Net variants in Table~\ref{tab:ablation}. The model C, combining SqueezeNet and 1D-LSTM, obtains a good balance between accuracy and memory print (6.6 MB). The best-performing variant G occupies memory of 68.8 MB, due to the larger dimensionality of hidden LSTM channels and the 2D
LSTM structure.

\section{Conclusions}
\label{sec:conclusions}

Our work introduces TCC-benchmark, by far, the largest temporal color constancy dataset of high resolution images. 
More than 20 popular methods were evaluated on the dataset
including the recent state-of-the-arts. As a new baseline
method, we proposed TCC-Net which is an end-to-end learnable deep and recurrent neural network architecture.
TCC-Net achieves state-of-the-art results on our
TCC-benchmark and SFU Gray Ball used in the previous
works on temporal CC. TCC-Net outperforms, in terms of mean angular error, the best single-image and temporal color constancy methods by 33\% and  30\%, respectively, on the SFU Gray Ball set  and by 40\% and  27\%, respectively, on the TCC-benchmark. 
The TCC-Net represents a technique for combining SqueezeNet and 2D-LSTM to capture spatial-temporal variations in a video. 
We present multiple variants of
TCC-Net including ones with small memory consumption and
therefore suitable for mobile devices.

\clearpage

\bibliographystyle{splncs}
\bibliography{color_constancy}
\end{document}